\documentclass[runningheads]{llncs}
\usepackage{graphicx}

\usepackage{tikz}
\usepackage{comment}
\usepackage{amsmath,amssymb} 
\usepackage{color}

\usepackage{xurl}

\begin{document}
\pagestyle{headings}
\mainmatter
\def\ECCVSubNumber{5673}  

\title{SemifreddoNets: Partially Frozen Neural Networks for~Efficient~Computer~Vision~Systems} 

\titlerunning{SemifreddoNets: Partially Frozen Neural Networks}
%

\author{Leo F Isikdogan, Bhavin V Nayak, Chyuan-Tyng Wu, Joao Peralta Moreira, Sushma Rao, and Gilad Michael}
\institute{Intel Corporation, Santa Clara, CA}

\authorrunning{Isikdogan et al.}
%
\maketitle

\begin{abstract}
We propose a system comprised of fixed-topology neural networks having partially frozen weights, named SemifreddoNets. SemifreddoNets work as fully-pipelined hardware blocks that are optimized to have an efficient hardware implementation. Those blocks freeze a certain portion of the parameters at every layer and replace the corresponding multipliers with fixed scalers. Fixing the weights reduces the silicon area, logic delay, and memory requirements, leading to significant savings in cost and power consumption. Unlike traditional layer-wise freezing approaches, SemifreddoNets make a profitable trade between the cost and flexibility by having some of the weights configurable at different scales and levels of abstraction in the model. Although fixing the topology and some of the weights somewhat limits the flexibility, we argue that the efficiency benefits of this strategy outweigh the advantages of a fully configurable model for many use cases. Furthermore, our system uses repeatable blocks, therefore it has the flexibility to adjust model complexity without requiring any hardware change. The hardware implementation of SemifreddoNets provides up to an order of magnitude reduction in silicon area and power consumption as compared to their equivalent implementation on a general-purpose accelerator.
\end{abstract}

\section{Introduction}
On-device artificial intelligence (AI) applications are becoming increasingly common for a wide variety of products, including smartphones, autonomous vehicles, drones, and different types of robots. Many, if not most, of those `visually intelligent' devices today are powered by convolutional neural networks that run either on cloud computing platforms or the device itself.

Cloud-based services rely on an internet connection to operate and transmit data back and forth between the device and the remote servers, which results in high latency. Therefore, they are typically not suitable for real-time applications. On-device systems, on the other hand, do not rely on remote resources, and therefore run with much less latency. Furthermore, on-device computing usually provides a higher level of security than cloud-based applications since the user data never needs to leave the device. However, running everything end-to-end on a low power device remains a challenging task, since many computer vision applications require a substantial amount of computing power to run in real-time. Therefore, on-device solutions may need expensive and large accelerators to achieve low latency and high throughput.

Many computer vision applications use custom, highly specialized convolutional neural network (CNN) architectures tailored for their target tasks. One way to reduce the complexity of a neural network inference hardware would be to fix the topology of a given network and implement it as a fixed-function style, in-line hardware block. Until recently, fixing the topology was not a feasible approach given the pace of development in network architecture design. For example, the top-5 error rate in the ImageNet Large Scale Visual Recognition Challenge~\cite{russakovsky2015imagenet} for the winning models went from $16.4\%$~\cite{krizhevsky2012imagenet} in 2012 to $6.67\%$~\cite{szegedy2015going} in 2014, and to $3.57\%$~\cite{he2016deep} in 2015. The complexity of the top-performing models also made them difficult to implement in fully-pipelined hardware. As the network topologies matured and more efficient neural network design patterns emerged~\cite{howard2017mobilenets,zhang2018shufflenet,sandler2018mobilenetv2,ma2018shufflenet}, hard-wiring at least a portion of a neural network topology~\cite{whatmough2019fixynn,wu2019trainable,visionisp,asama2020processing,asama2020machine} became a somewhat less flexible but more efficient alternative to doing all the computation on general-purpose CNN accelerators.

\begin{figure}[t!]
\centering
\vspace{10pt}
\includegraphics[width=0.7\linewidth]{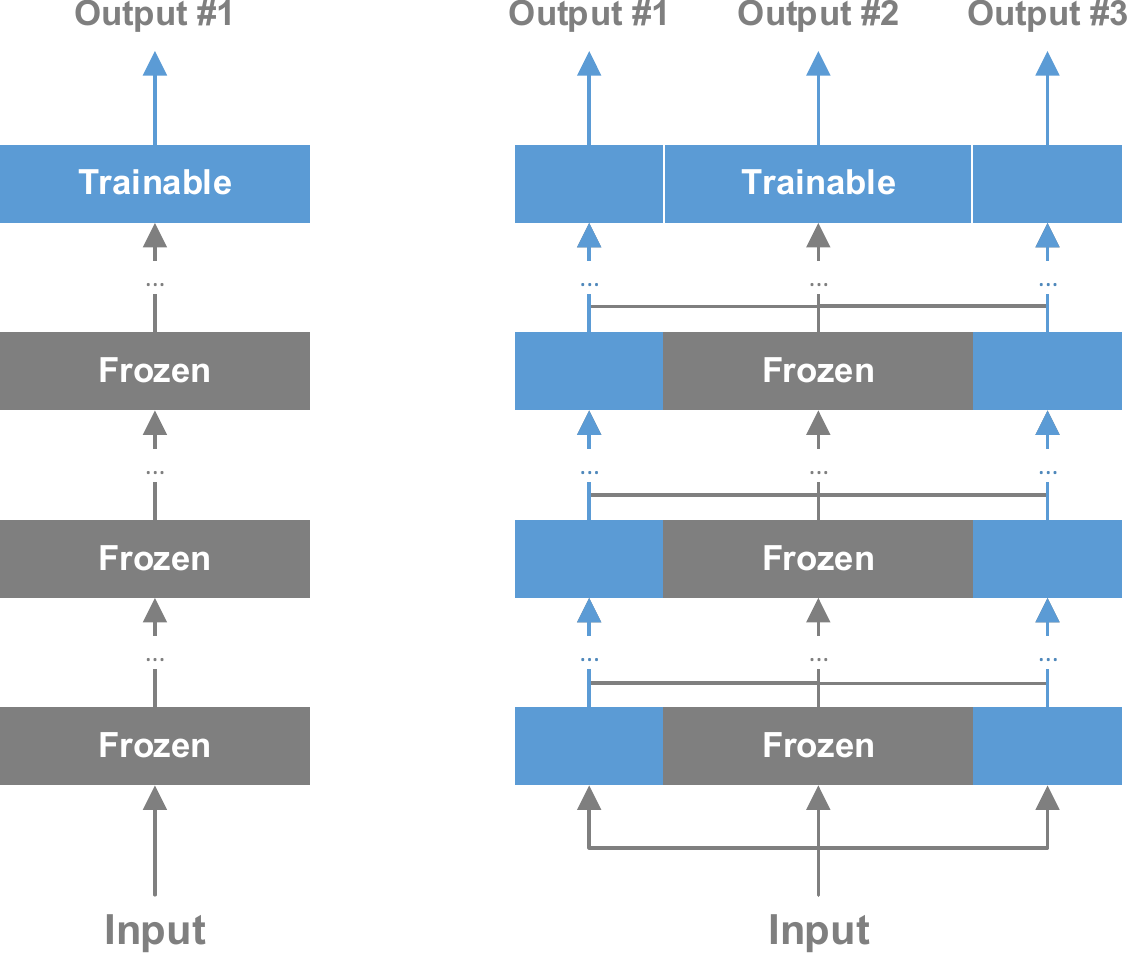}
\vspace{5pt}
\caption{A high-level illustration of how the vertical freezing scheme in SemifreddoNets (right) differs from traditional layer-level parameter freezing approaches (left): the grey blocks indicate the frozen weights whereas the blue blocks show the parts having trainable variables. SemifreddoNets provide some room for adaptation at both lower-level and higher-level feature extractors, whereas the traditional approach freezes specific level layers entirely.}
\label{fig:semifreddo_intro}
\end{figure}

Using a fixed-topology model relies on the idea that a model that works well for one task is likely to generalize for other similar types of problems. Although searching for a custom network architecture for each task is shown to have some value~\cite{zoph2016neural}, we argue that the efficiency benefits of using a fixed-topology model outweigh the marginal value of application-specific topologies. We propose a solution that significantly reduces the hardware complexity by using a fixed-topology neural network and partially frozen weights. We named this architecture after the Italian dessert, Semifreddo, due to its semi-frozen feature extractors. The frozen part is fixed in hardware and is designed to generalize across different tasks and input data types. The trainable part consists of configurable weights across varying levels of abstraction, leaving room for both adapting to new tasks and new kinds of data.

We optimize our system to work on very low power environments to bring significant AI capabilities to almost any consumer device. Our proposed hardware consists of three interconnected cores that run up to three different tasks at a time, processing up to 200 frames per second on VGA input. Our hardware is fully pipelined and does not use time multiplexing unlike conventional CNN accelerators in the single pass mode. Therefore, the pipeline runs at a constant 100\% steady state utilization, whereas conventional, fully-programmable CNN accelerators typically operate at around 40\% utilization rate. All three cores use a combined silicon area of 4mm$^2$ modeled with TSMC 16nm technology. Our fully-pipelined, partially-frozen architecture leads to savings in silicon area by a factor of $\sim$4$\times$ to $10\times$ as compared to generic CNN accelerators that have a hardware footprint of $\sim$15mm$^2$.

\section{Related Work}
Many CNN accelerators in the market provide hardware acceleration for computer vision applications, including Intel Movidius Vision Processing Units~\cite{movidius}, Google Coral Edge Tensor Processing Units~\cite{google-coral}, and Nvidia Jetson modules~\cite{nvidia-jetson}. Although our system can replace those accelerators in many use cases, we did not design it to be a general-purpose neural network accelerator. Our hardware targets ultra-low-power systems having minimal silicon area budgets, where using a fully-blown CNN accelerator would not be feasible.

Our work resembles image signal processing hardware accelerators in the sense that it applies a series of filters to a given image. The closest work to ours is the recently published FixyNN hardware by Whatmough et al.~\cite{whatmough2019fixynn}, which used a fixed feature extractor that froze the first-N layers of a given model and did the rest of the computation on a generic deep learning accelerator. Although freezing the first layers increases the hardware efficiency, it does not leave much room for domain adaptation. For example, if the frozen parameters were pre-trained on RGB images, a fixed feature extractor would not be able to fully utilize different types of inputs, such as depth maps or feature maps extracted by other networks~\cite{visionisp}. Furthermore, using a fully-programmable head would decrease the overall efficiency of the system, particularly under very low silicon area budgets, where the programmable head would significantly bottleneck the fixed feature extractor.

Our solution leaves some of the weights trainable at different levels of depth in the model, rather than freezing the layers entirely and doing the rest of the computation on a programmable CNN accelerator. Our model is implemented as an in-line hardware block as a whole from inputs to outputs. It is possible to scale our model linearly to any area budget at 100\% utilization rate without bottlenecking the efficiency at specific parts of the model.

Our work also bears similarities with multi-task learning methods that share weights~\cite{guo2019spottune,misra2016cross,mallya2018piggyback}. However, to the best of our knowledge, no solutions available in the market combine fixed and trainable weights side-by-side across varying levels of feature complexity, for the purpose of silicon area minimization.

\section{SemifreddoNets}
Implementing a deep convolutional neural network in fully-pipelined hardware provides numerous benefits over using general-purpose accelerators. For example, fixed-function-style neural network hardware can reach a utilization rate of 100\% as compared to 40\% typical utilization rate in generic CNN accelerators. However, building such hardware has been challenging due to the sheer number of parameters that many modern CNNs have. Those parameters cost significant silicon area when the weights are stored in dedicated memory. Indeed, time multiplexing of hardware accelerators could help reduce the memory requirement. However, this would also decrease the overall efficiency of the system. A highly-efficient, fully-pipelined neural network hardware would require all weights to be kept in memory simultaneously. The high cost associated with the weights makes a fully trainable model not feasible for small area budgets.

We address this problem by fixing and hard-wiring some of the weights in our model. For the fixed weights, we use fixed scalers with a single input to substitute the corresponding multipliers. This approach not only saves the memory that would store the parameters but also reduces the complexity of the logic design by replacing the multipliers with cheaper scalers and pruning zero weights. We store the remaining weights in SRAM and leave them configurable to retain an ample amount of flexibility in the model.

Our hardware represents all weights and intermediate feature maps as 8-bit signed fixed-point numbers. To ensure quantization-friendly values in feature maps, we prevent the model from producing zero-variance feature maps during training. We automatically detect neurons leading to zero-variance outputs by monitoring the moving variance parameter in the batch normalization layers. At the end of each epoch, we re-initialize the weights corresponding to very small moving variance values. This process mainly detects and resuscitates nearly-dead neurons during training. Our approach is somewhat similar to the recently published neural rejuvenation~\cite{qiao2019neural} approach, which aimed to identify and re-initialize dead neurons for better resource utilization.

\subsection{Vertically Frozen Neural Networks}
What parts of the model to freeze is an important design choice that can impact the behavior and capabilities of the model. In the literature, it is a common practice to freeze the first N-layers of a neural network as a form of transfer learning~\cite{yosinski2014transferable}. This type of parameter freezing is usually done to speed up training and to reduce the risk of overfitting. Similarly, it is also possible to train only the first layers while keeping the rest of the network frozen to adapt an already trained model to different input data. Freezing the first layers would work well on similar input data, whereas freezing the last layers would generalize well for similar tasks.

\begin{figure}[t]
\centering
\vspace{5pt}
\includegraphics[width=0.85\linewidth]{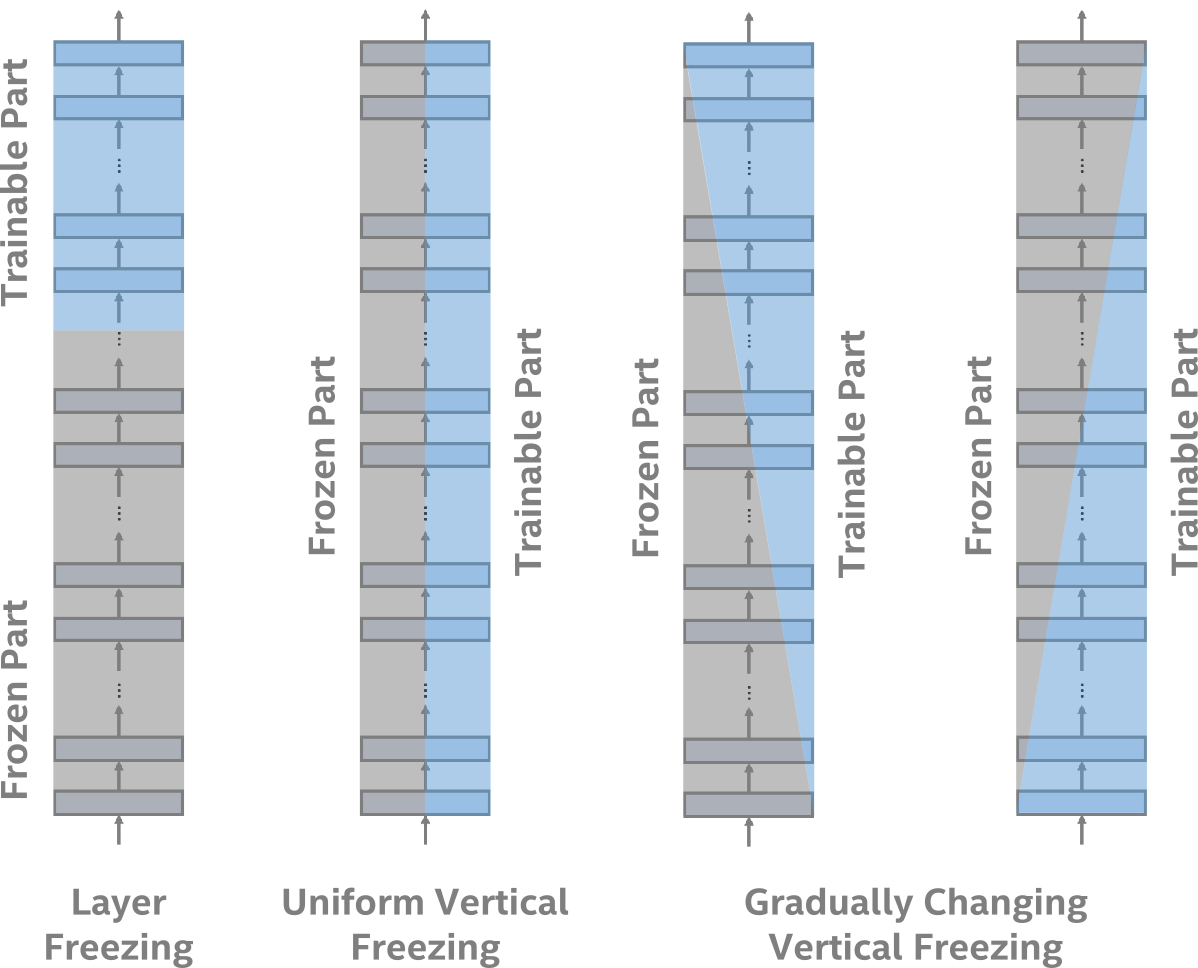}
\vspace{5pt}
\caption{An illustration of how vertical weight freezing schemes differ from layer-level freezing approaches. Vertical weight freezing has the flexibility to both adjust to different types of input data and tasks. Many different types of vertical weight freezing schemes can be tailored to different kinds of needs. For example, if the system is expected to perform various tasks while the input remains the same, then the freezing ratio can be decreased gradually from the input layer to the output layer. Similarly, if the system is expected to perform similar tasks, but the input data source may vary, then the freezing ratio can be increased gradually.}
\label{fig:freezing_schedules}
\end{figure}

We propose a more balanced parameter freezing scheme that has the flexibility to both adjust to different types of input data and tasks. Unlike traditional layer-level horizontal freezing approaches, our method vertically freezes a portion of the weights, distributed across the layers (Fig.~\ref{fig:semifreddo_intro}). The proportion of frozen weights can be uniform across the layers as well as changing gradually, depending on the needs (Fig.~\ref{fig:freezing_schedules}). A very basic weight freezing scheme would fix a certain percentage of all weights in each layer, where the silicon area budget determines the freezing rate. This approach would essentially create a slice of trainable variables in the model.

We can take this idea one step further and create multiple trainable slices that share the same frozen parts. Those trainable slices can either be used to perform different tasks on the same input or act as one large network to perform one task with higher accuracy. Based on this idea of vertical weight freezing, we propose a neural network architecture, named SemifreddoNets, that achieves a small hardware footprint, low power, low cost, and high efficiency.

\subsection{Backbone Model Architecture}
\begin{figure}[t]
\centering
\vspace{5pt}
\includegraphics[width=0.7\linewidth]{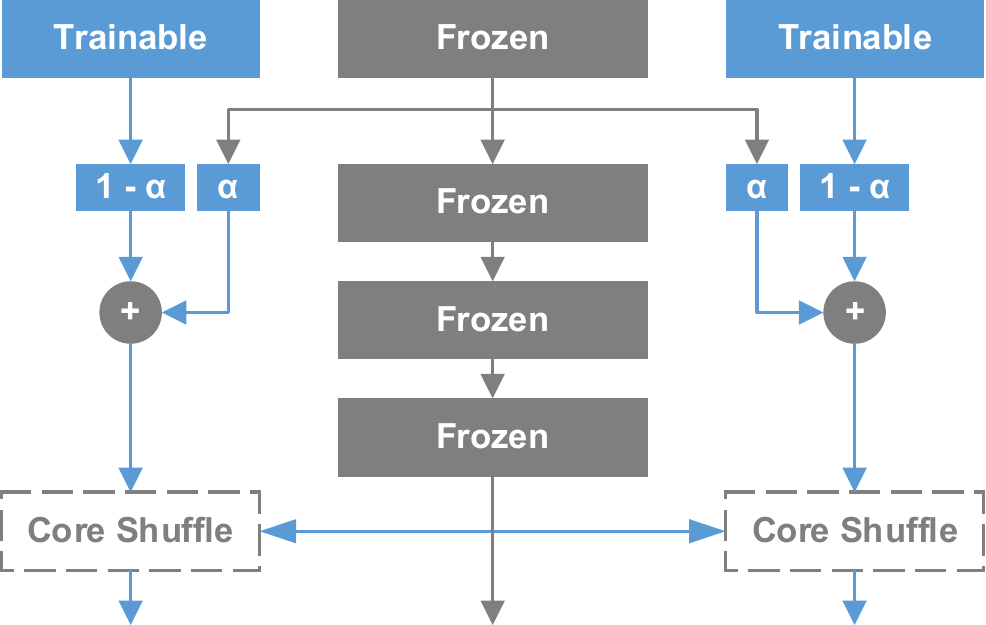}
\vspace{10pt}
\caption{A Semifreddo module consists of one frozen and two trainable cores (trainable parts shown in blue). The trainable cores selectively transfer features from the frozen core using trainable alpha blending parameters. The modular architecture allows for both using each core independently, to perform different tasks, and in conjunction with each other to perform a single task with higher representational power. The optional core shuffle module lets the two trainable cores exchange feature maps when both cores are trained to do the same task.}
\label{fig:semifreddo_module}
\end{figure}

The macro architecture of our system consists of one frozen core and two parallel trainable cores (Fig.~\ref{fig:semifreddo_intro}), where the trainable cores have fewer layers and therefore are smaller. Both the frozen and trainable cores have their topology hard-wired. The frozen core is trained once, whereas the trainable parts are trained separately for each given dataset and task. In our experiments, we trained the frozen core on ImageNet data since CNNs pre-trained on this dataset typically learn useful general-purpose features~\cite{huh2016makes}. Before the weights are fixed in hardware, the frozen core can also be trained on other datasets in a multi-task setting depending on the needs. The frozen core aims to provide features that are general-purpose enough for the target applications. The trainable cores selectively transfer and enrich those features using trainable alpha blending parameters (Fig.~\ref{fig:semifreddo_module}).

We define a trainable alpha blending layer as
\begin{equation}
    \begin{split}
    \alpha =& ~\sigma(w) \\
    y =& ~\alpha\cdot x_f + (1 - \alpha) \cdot x_t
    \end{split}
\end{equation}
for each input channel, where $w$ is a randomly initialized trainable parameter, $\sigma$ is the sigmoid function, and $x_f$ and $x_t$ are the outputs of the frozen and trainable blocks in the preceding layer, respectively. The alpha blending layer acts as a gating mechanism between the cores and helps the model decide the strength of transfer learning on a feature map basis (Fig.~\ref{fig:semifreddo_module}). Although the alpha parameters are learned during training, they can also be manually set to a particular value to enforce certain behavior. For example, setting all alpha parameters to zero would separate all three cores by disabling the data flow between the cores entirely. Similarly, setting them all to 0.5 would turn the trainable cores into residual feature extractors.

All three cores act as backbone networks that feed feature maps to application-specific model heads for up to three different tasks at a time. The cores can run both independently and together with each other. For example, one can use the frozen core output for image classification, one of the trainable cores for scene classification, and the other trainable core for semantic segmentation.

\begin{figure}[t]
\centering
\vspace{5pt}
\includegraphics[width=0.7\linewidth]{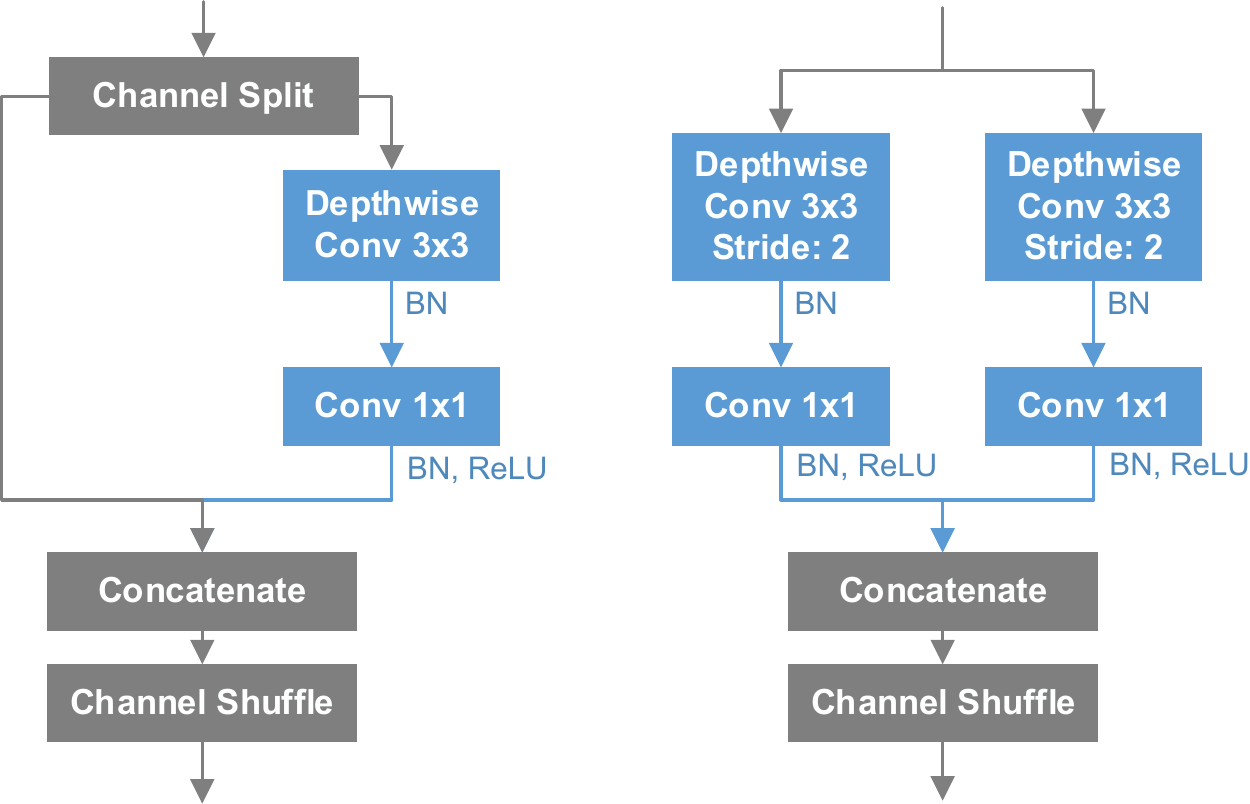}
\vspace{10pt}
\caption{We use simplified versions of ShuffleNetV2~\cite{ma2018shufflenet} blocks to implement our SemifreddoNet cores: regular building blocks (left) and downscaling blocks (right).}
\label{fig:shuffle_blocks}
\end{figure}

Any efficient neural network architecture can be used to implement our SemifreddoNet macro architecture. We used a network topology based on ShuffleNetV2~\cite{ma2018shufflenet} to implement the building blocks of our model. Each block in the system consists of a channel split, followed by depthwise separable convolution, channel concatenation, and uniform channel shuffle (Fig.~\ref{fig:shuffle_blocks}). The blocks that downsample their inputs skip the channel split operator and use a stride of two in the depthwise convolution. Therefore, they double the number of channels while reducing the feature map size by a factor of two in both horizontal and vertical axes. When two trainable cores are used for one task, we also shuffle feature maps between trainable cores by swapping half of the feature maps at the output of each alpha blending layer (Fig.~\ref{fig:semifreddo_module}). This cross-core channel shuffling helps both cores act as a single network more efficiently.

The alpha blending layers between the frozen and trainable cores require the shape of the input feature maps to match. Therefore, all cores have intermediate feature maps that match in size. The trainable cores are made smaller by carving out some of the repeated layers rather than reducing the number of trainable kernels per layer while keeping both cores in synch with each other in the pipeline. We call each one of these building blocks that consist of parallel trainable and frozen layers, a Semifreddo module (Fig.~\ref{fig:semifreddo_module}). Our model architecture consists of repeated blocks of Semifreddo modules. The overall architecture breakdown is shown in Table \ref{tab:model_architecture_breakdown}.

\begin{table}[t]
\centering
\vspace{5pt}
\includegraphics[width=0.8\linewidth]{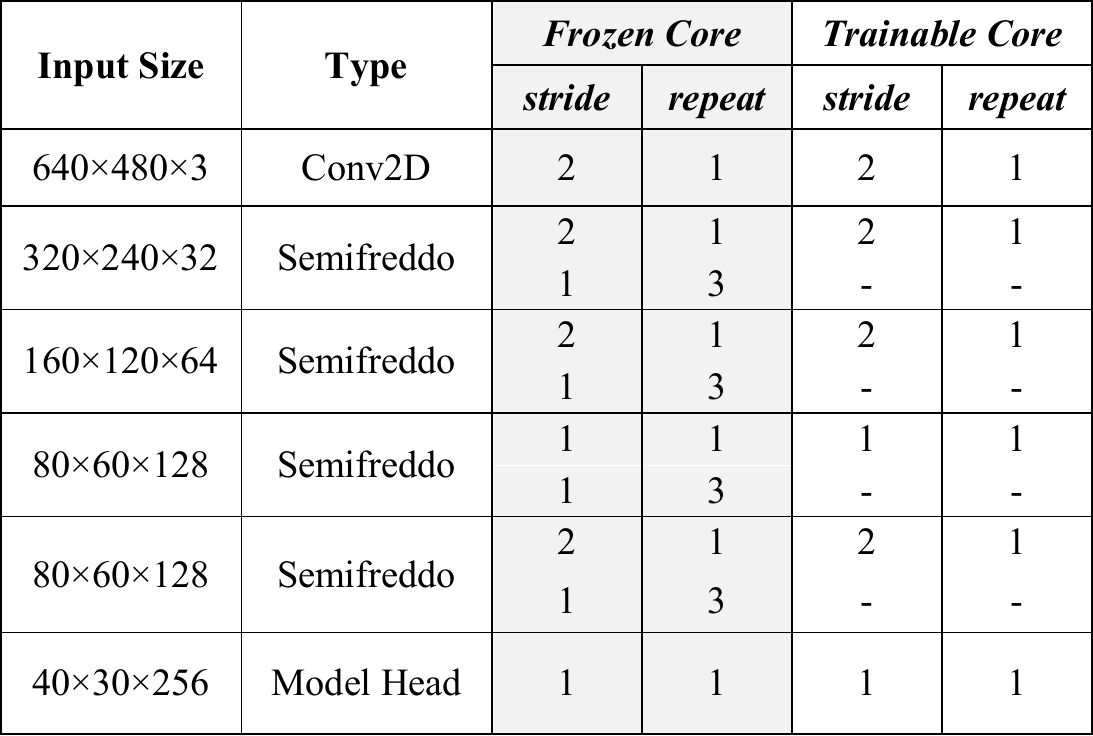}
\vspace{5pt}
\caption{Architectural breakdown of the frozen and trainable cores: the trainable cores have a smaller number of repeated blocks, therefore have fewer layers. Both trainable and frozen cores have intermediate feature maps that match in size.}
\label{tab:model_architecture_breakdown}
\end{table}

We use the Semifreddo modules to freeze a model vertically. Freezing the parameters this way (Table~\ref{tab:results}) produced comparable results to freezing a certain percentage of parameters in each layer uniformly (Fig.~\ref{fig:freeze_ratio}), while providing additional benefits. One advantage of using Semifreddo modules instead of fully-uniform freezing is the ease of implementation. For example, we needed to modify our code at the optimizer level to implement uniform weight freezing. On the other hand, the frozen and trainable parts in Semifreddo modules can easily be defined in any mainstream deep learning framework and trained without modifying the parameter update mechanisms in the underlying framework. Another advantage of using Semifreddo modules to freeze a model vertically is the ability to decouple the frozen and trainable cores. This modular architecture allows for training the trainable and frozen cores separately for different tasks.

\subsection{Model Head}
The backbone model in our system outputs feature maps that need to be further processed to perform a computer vision task. Those feature maps can be used as-is in a host system that has additional computing capabilities, such as having a digital signal processor (DSP). However, a host system might not always have such additional hardware to process the raw feature maps.

To build a standalone system, we propose a multi-purpose model head block that can perform basic computer vision tasks without relying on the compute capabilities of a host system. The model head inputs feature maps and produces output vectors for a given task.

We implement this model head as a pointwise convolution layer having a configurable number of outputs. The model head supports up to 131072 parameters, which would be sufficient for many types of basic computer vision tasks. For example, given 256-channel feature maps from each trainable core, the model head would be able to classify up to 256 kinds of scenes and segment up to 256 types of objects simultaneously. The head supports group convolutions to handle larger outputs while staying within the limits of the total number of configurable weights.

The head implements an optional pooling operator and a configurable activation function that can be enabled when needed. The pooling operator is enabled when the entire image needs to be analyzed to make a single prediction, such as image classification and face identification and is disabled for the tasks that require spatial granularity. The model has a total downscaling factor of 16$\times$ when the pooling is disabled. The global average pooling operator runs as a running-average accumulator, as the images are acquired line-by-line in the raster scan order. Finally, the activation function at the end of the model head is designed to approximate any arbitrary activation function as a piecewise linear function.

We designed this hardware model head to run basic computer vision tasks on simple devices that do not have any additional compute capabilities. For more sophisticated tasks, we also provide the option of outputting the feature maps and implementing more complex neural network heads on the host device.

\subsection{Repeatable Blocks}
\label{sec:repeatable}
Fixing the model topology helped us design highly efficient neural network hardware, while somewhat limiting the flexibility of our models. As different tasks may need models having varying levels of capacities, we propose a modular design scheme to adjust the model depth without duplicating the logic in the hardware. Our modular design scheme implements deeper and larger network architectures by cycling the feature maps over the same hardware blocks.

\begin{figure}[t]
\centering
\vspace{5pt}
\includegraphics[width=1.0\linewidth]{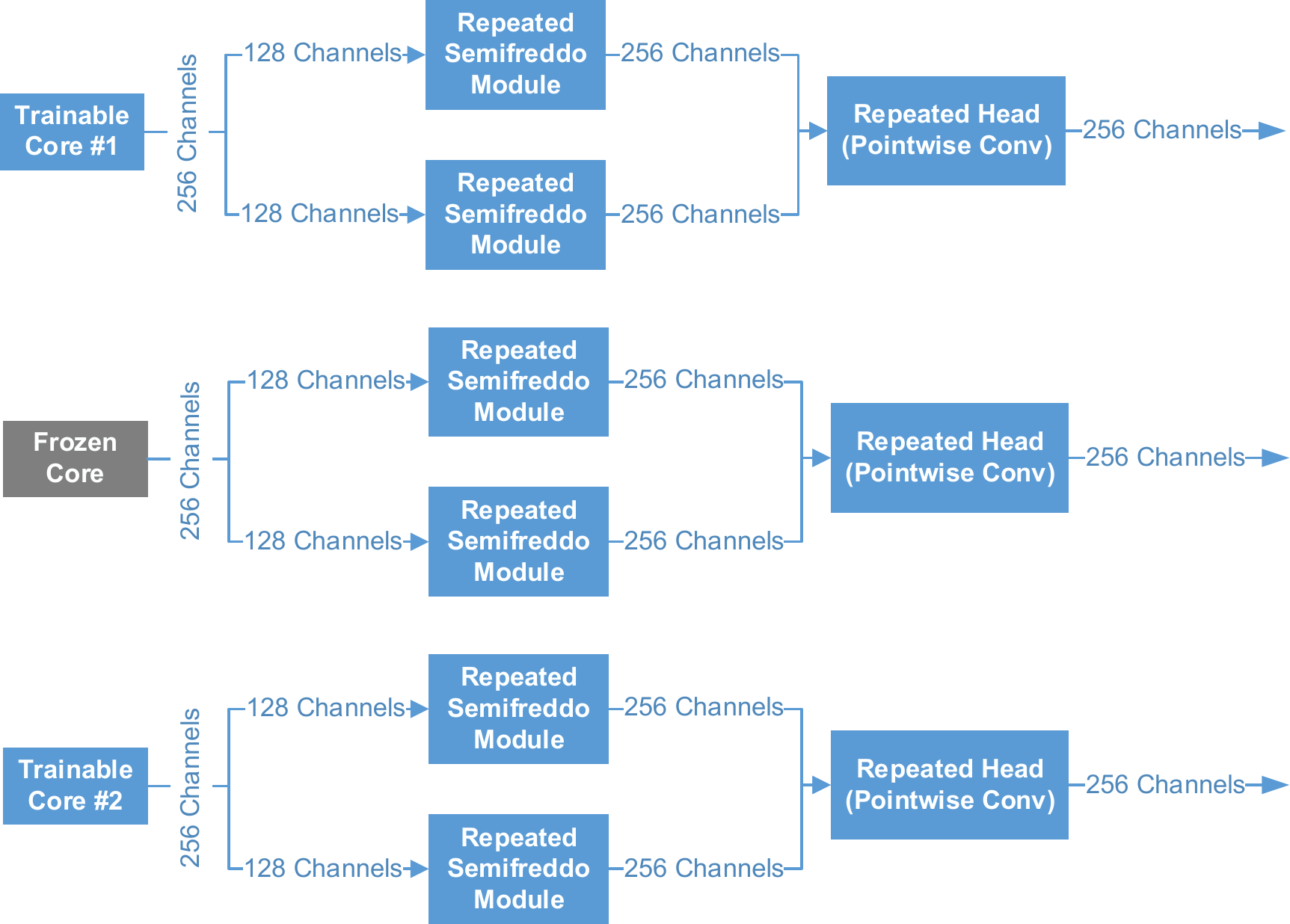}
\vspace{10pt}
\caption{Repeatable blocks: the last blocks in our system can be repeated to increase the capacity of the underlying models. The Semifreddo modules can be repeated as many times as necessary to meet a given accuracy requirement, at the expense of inference speed.}
\label{fig:modular}
\end{figure}

In particular, we modularize the last trainable Semifreddo blocks and the model head. We reuse them repeatedly in a single inference pass to improve model accuracy when needed (Fig.~\ref{fig:modular}). We can repeat the modular blocks as many times as necessary. However, reusing the building blocks for different layers requires the weights to be reloaded every time an existing hardware block is used in place of a new one. Therefore, implementing larger models this way comes at the cost of lower inference speeds. Nevertheless, the block modularity provides the flexibility to find a reasonable balance between accuracy and speed (Table \ref{tab:repeated_blocks}), given a set of requirements.

\begin{table}[t]
\centering
\vspace{5pt}
\includegraphics[width=0.45\linewidth]{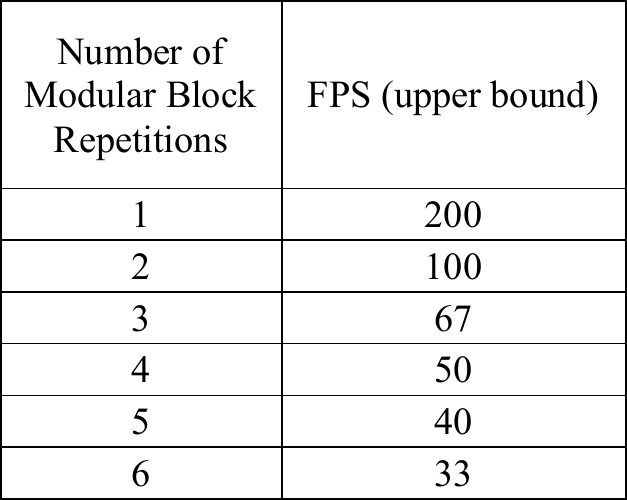}
\vspace{5pt}
\caption{Impact of repeatable hardware blocks on inference speed in terms of maximum number of frames per second that can be processed at VGA resolution.}
\label{tab:repeated_blocks}
\end{table}

\section{Results}
To evaluate the value of the frozen features and the trainable cores, we performed a set of experiments covering different configurations of our model on different types of tasks (Table~\ref{tab:results}). We used two computer vision tasks that have significantly different types of input data: image classification and face identification.

\begin{table}[t]
\centering
\vspace{5pt}
\includegraphics[width=1.0\linewidth]{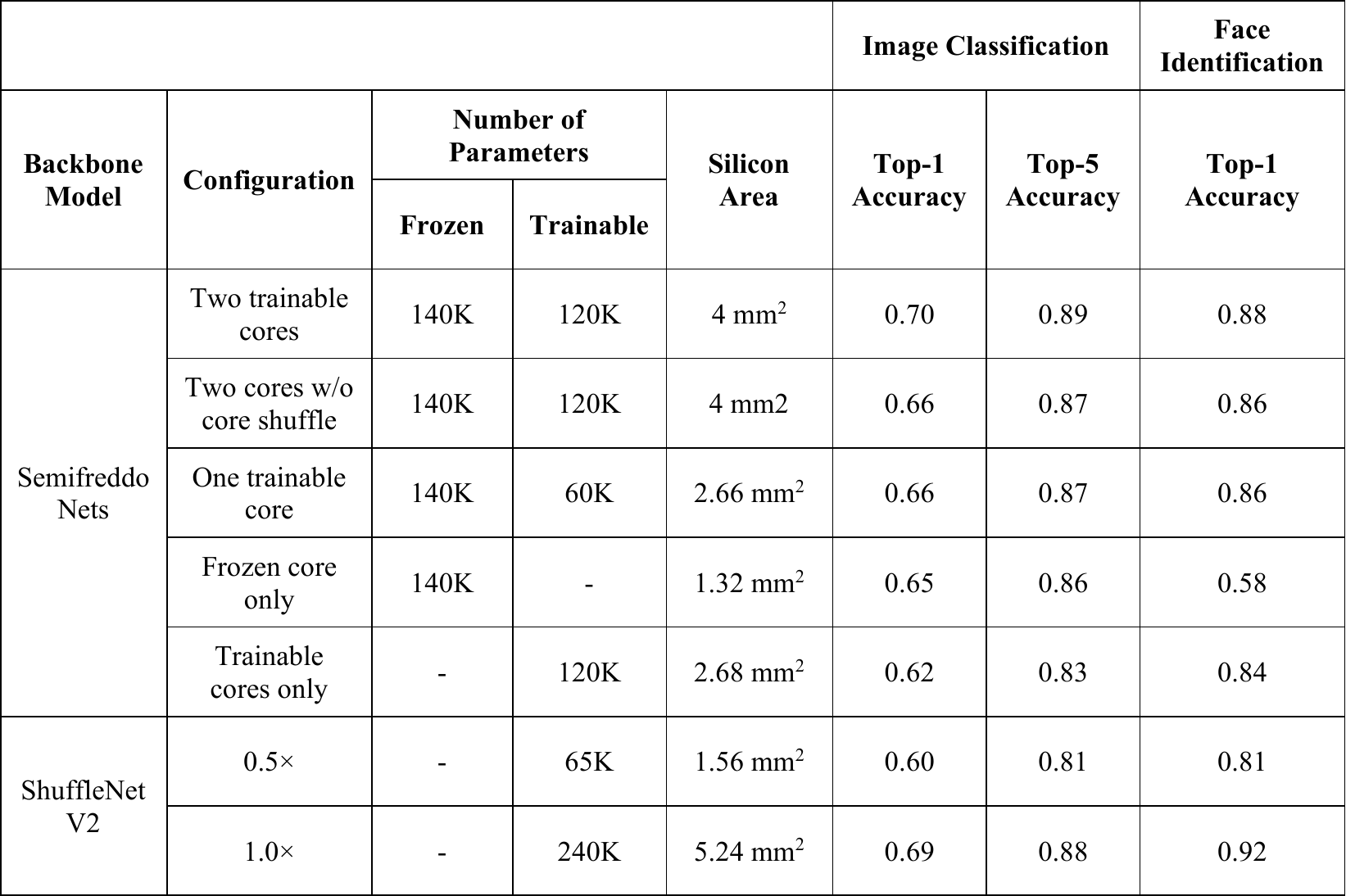}
\vspace{5pt}
\caption{Comparison of different configurations of SemifreddoNets and ShuffleNetV2 in terms of the number of parameters, silicon area, and performance on various computer vision tasks. The number of parameters and silicon area exclude the model head. For image classification, all backbone networks use the same model head that the original ShuffleNetV2 used. Face classification models use our proposed lightweight model head. The frozen core was pre-trained on ImageNet; therefore, its additional value was higher for image classification than for face classification. The frozen core can also be trained in a multi-task setting to maximize its value for a given set of tasks.}
\label{tab:results}
\end{table}

The first task, image classification on the ImageNet challenge dataset, used a training setup identical to the frozen core pretraining. Therefore, it was expected to benefit from the frozen core the most. The second task, face identification, used the VGGFace2~\cite{cao2018vggface2} and LWF~\cite{huang2008labeled} face datasets for training and test, respectively. Both of those datasets had a data distribution that is significantly different from ImageNet. We used the training setup in~\cite{schroff2015facenet} as-is, without any further hyperparameter tuning.

As a benchmark, we used fully trainable ShuffleNetV2 backbones on the same tasks. In ShuffleNetV2 models, we used width multipliers of 0.5 and 1.0 to get backbone networks that are closest to our models in terms of the hardware footprint and the total number of parameters. We used the same training setup as the original ShuffleNetV2 paper~\cite{ma2018shufflenet} for both our SemifreddoNet backbones and the benchmark models.

In the image classification experiments, we used the same model head that the original ShuffleNetV2 paper used. Since the model head would be too large to run on our proposed in-line model head block, we assumed that the head would run on a DSP. In the face identification experiments, we used our proposed lightweight model head, which does not rely on any additional hardware on the host system.

As expected, the value of frozen features were higher for image classification than for face identification. However, the face identification task still benefited from the ImageNet-trained frozen core, despite the differences in input data distributions. Overall, SemifreddoNets performed comparably or better than their fully-trainable ShuffleNetV2 counterparts, given small silicon area budgets.

\section{Ablation Study}
\begin{figure}[t]
\centering
\includegraphics[width=0.95\linewidth]{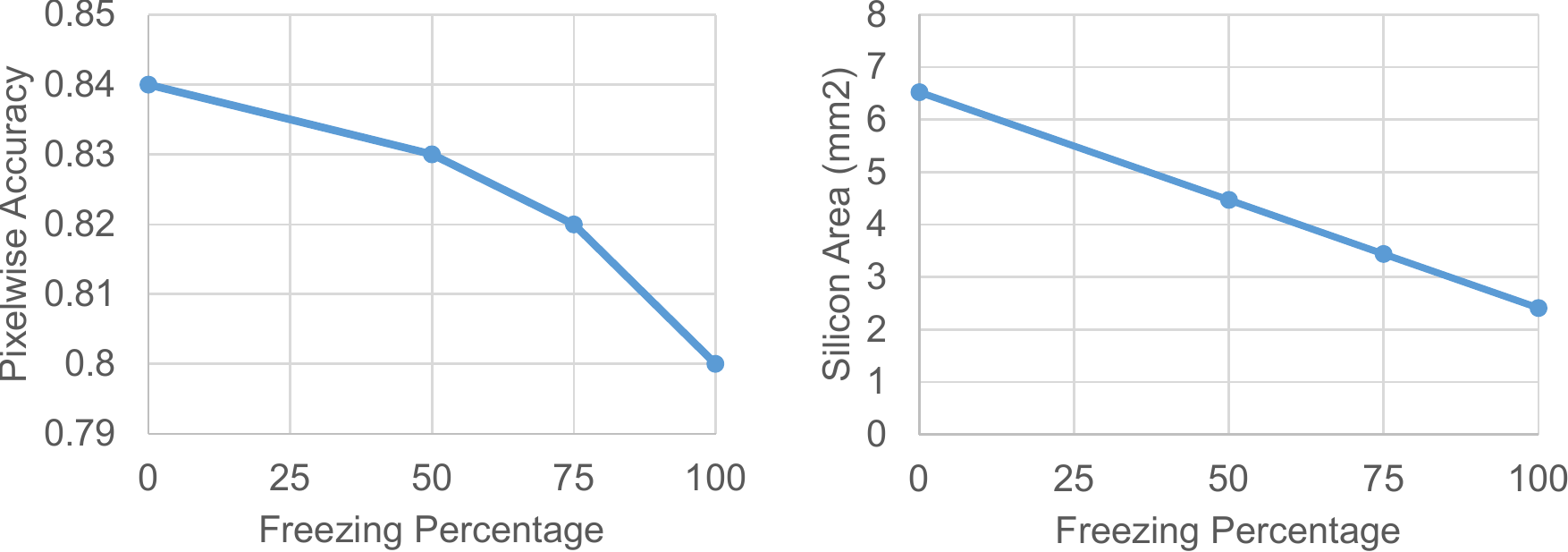}
\caption{Impact of weight freezing ratio on silicon area and pixelwise accuracy for an exemplary, semantic segmentation task. Freezing a larger portion of the parameters reduces the hardware footprint, however it also decreases the accuracy of the model.}
\label{fig:freeze_ratio}
\end{figure}

We performed a set of experiments to measure the impact of our design choices in our model architecture.

\vspace{5pt}
\noindent\textbf{Proportion of frozen weights.} We first tested the impact of vertical weight freezing on the flexibility on the model. We started with a backbone model that is twice as wide as the frozen core in our final network architecture. We pretrained this backbone model on ImageNet. Then, we uniformly froze 0\%, 50\%, 75\%, and 100\% of the parameters in the backbone, while leaving the model head trainable. We fine-tuned remaining weights in the backbone and the model head on the Citycapes~\cite{cordts2016cityscapes} dataset to perform pixelwise semantic segmentation. As expected, having a larger portion of the network frozen resulted in lower accuracy. The largest performance drop occurred between the 75\% and 100\% freezing ratios (Fig.~\ref{fig:freeze_ratio}). SemifreddoNets do not support specifying an exact freezing ratio as they are not frozen in an entirely uniform manner, unlike in this experiment. However, we can still approximate a given freezing ratio. The effective freezing ratio in our final backbone model is 77\% for using a single trainable core and 54\% for using both of the trainable cores. Those freezing ratios provided a good trade-off between accuracy and silicon area.

\vspace{5pt}
\noindent\textbf{Repeatable blocks.}
We measured the impact of repeatable hardware blocks on the performance by training models having different numbers of repeated Semifreddo blocks. Repeating the last Semifreddo blocks twice, as shown in Fig.~\ref{fig:modular}, increased the face identification accuracy from 88\% to 96\%. However, the block repetitions also increased the overall delay, reducing the frames per second that the system can process (Table~\ref{tab:repeated_blocks}). Repeating the blocks further led to only minor further improvements in the accuracy (up to 2\%).

\vspace{5pt}
\noindent\textbf{Cross-core shuffle.} Shuffling the feature maps between the two trainable cores helped both cores act as a single, larger network. Using both trainable cores for the same task improved the performance only when the core shuffling was enabled. Without the core shuffle, the additional trainable core led to no significant gains in the performance metrics (Table~\ref{tab:results}). Core shuffling improved the results while having a negligible cost in hardware.

\vspace{5pt}
\noindent\textbf{Pointwise convolutions.} ShuffleNet V2 blocks use pointwise convolutions followed by depthwise and pointwise convolutions. To save silicon area, we dropped the first pointwise convolutions in our backbone network (Fig.~\ref{fig:shuffle_blocks}). Dropping the first convolutions in each branch made no difference in accuracy in the first two decimal places for face identification task, and a 2\% absolute drop in top-5 accuracy for the image classification task.

\vspace{5pt}
\noindent\textbf{Trainable batch normalization parameters.} In the frozen core, we left the batch normalization parameters trainable to help adapt the frozen feature extractor to different types of inputs. We observed an absolute 2\% drop in accuracy for the face identification task, when the batch norm parameters in the frozen core were not fine tuned. Our results confirmed the findings of previous studies~\cite{whatmough2019fixynn,chen2017rethinking}, which showed the impact of trainable batch normalization parameters on transfer learning.

\section{Conclusions}
We proposed fixed-topology neural network blocks that vertically froze network parameters for hardware efficiency. Our proposed weight freezing scheme significantly reduced the hardware footprint while maintaining a fair amount of flexibility. Our proposed system has a modular architecture that is straightforward to use, extend, and integrate into existing systems. We demonstrated the capabilities of an exemplary neural network hardware architecture that consisted of one frozen and two trainable cores. Our work can potentially be extended to have more trainable cores. Increasing the number of trainable cores would decrease the marginal cost of the frozen core, allowing for building high-performance inference engines that can handle many more tasks at a time.


\begin{thebibliography}{10}
\providecommand{\url}[1]{\texttt{#1}}
\providecommand{\urlprefix}{URL }
\providecommand{\doi}[1]{https://doi.org/#1}

\bibitem{google-coral}
{Coral.ai}. \url{https://coral.ai/products/}

\bibitem{movidius}
{Intel Movidius VPUs}.
  \url{https://www.intel.com/content/www/us/en/artificial-intelligence/movidius-myriad-vpus.html}

\bibitem{nvidia-jetson}
{Nvidia Jetson}.
  \url{https://www.nvidia.com/en-us/autonomous-machines/jetson-store/}

\bibitem{asama2020processing}
Asama, M., Isikdogan, F., Rao, S., Kalderon, A., Wu, C.T., Nayak, B., Moreira,
  J.P., Kounitsky, P., Berlin, B., Michael, G., et~al.: Processing images using
  hybrid infinite impulse response {(IIR)} and finite impulse response {(FIR)}
  convolution block (2020), {US Patent App.} 16/674,512

\bibitem{asama2020machine}
Asama, M., Isikdogan, L.F., Rao, S., Nayak, B.V., Michael, G.: A machine
  learning imaging core using separable {FIR-IIR} filters. arXiv preprint
  arXiv:2001.00630  (2020)

\bibitem{cao2018vggface2}
Cao, Q., Shen, L., Xie, W., Parkhi, O.M., Zisserman, A.: {VGGFace2}: A dataset
  for recognising faces across pose and age. In: IEEE International Conference
  on Automatic Face \& Gesture Recognition (FG 2018). pp. 67--74. IEEE (2018)

\bibitem{chen2017rethinking}
Chen, L.C., Papandreou, G., Schroff, F., Adam, H.: Rethinking atrous
  convolution for semantic image segmentation. arXiv preprint arXiv:1706.05587
  (2017)

\bibitem{cordts2016cityscapes}
Cordts, M., Omran, M., Ramos, S., Rehfeld, T., Enzweiler, M., Benenson, R.,
  Franke, U., Roth, S., Schiele, B.: The cityscapes dataset for semantic urban
  scene understanding. In: Proceedings of the IEEE Conference on Computer
  Vision and Pattern Recognition. pp. 3213--3223 (2016)

\bibitem{guo2019spottune}
Guo, Y., Shi, H., Kumar, A., Grauman, K., Rosing, T., Feris, R.: Spottune:
  transfer learning through adaptive fine-tuning. In: Proceedings of the IEEE
  Conference on Computer Vision and Pattern Recognition. pp. 4805--4814 (2019)

\bibitem{he2016deep}
He, K., Zhang, X., Ren, S., Sun, J.: Deep residual learning for image
  recognition. In: Proceedings of the IEEE Conference on Computer Vision and
  Pattern Recognition. pp. 770--778 (2016)

\bibitem{howard2017mobilenets}
Howard, A.G., Zhu, M., Chen, B., Kalenichenko, D., Wang, W., Weyand, T.,
  Andreetto, M., Adam, H.: {MobileNets}: Efficient convolutional neural
  networks for mobile vision applications. arXiv preprint arXiv:1704.04861
  (2017)

\bibitem{huang2008labeled}
Huang, G.B., Ramesh, M., Berg, T., Learned-Miller, E.: Labeled faces in the
  wild: A database for studying face recognition in unconstrained environments.
  Tech. Rep. 07-49, University of Massachusetts, Amherst (October 2007)

\bibitem{huh2016makes}
Huh, M., Agrawal, P., Efros, A.A.: What makes imagenet good for transfer
  learning? arXiv preprint arXiv:1608.08614  (2016)

\bibitem{krizhevsky2012imagenet}
Krizhevsky, A., Sutskever, I., Hinton, G.E.: Imagenet classification with deep
  convolutional neural networks. In: Advances in Neural Information Processing
  Systems. pp. 1097--1105 (2012)

\bibitem{ma2018shufflenet}
Ma, N., Zhang, X., Zheng, H.T., Sun, J.: {ShuffleNet V2}: Practical guidelines
  for efficient {CNN} architecture design. In: Proceedings of the European
  Conference on Computer Vision. pp. 116--131 (2018)

\bibitem{mallya2018piggyback}
Mallya, A., Davis, D., Lazebnik, S.: Piggyback: Adapting a single network to
  multiple tasks by learning to mask weights. In: Proceedings of the European
  Conference on Computer Vision. pp. 67--82 (2018)

\bibitem{misra2016cross}
Misra, I., Shrivastava, A., Gupta, A., Hebert, M.: Cross-stitch networks for
  multi-task learning. In: Proceedings of the IEEE Conference on Computer
  Vision and Pattern Recognition. pp. 3994--4003 (2016)

\bibitem{qiao2019neural}
Qiao, S., Lin, Z., Zhang, J., Yuille, A.L.: Neural rejuvenation: Improving deep
  network training by enhancing computational resource utilization. In:
  Proceedings of the IEEE Conference on Computer Vision and Pattern
  Recognition. pp. 61--71 (2019)

\bibitem{russakovsky2015imagenet}
Russakovsky, O., Deng, J., Su, H., Krause, J., Satheesh, S., Ma, S., Huang, Z.,
  Karpathy, A., Khosla, A., Bernstein, M., et~al.: Imagenet large scale visual
  recognition challenge. International Journal of Computer Vision
  \textbf{115}(3),  211--252 (2015)

\bibitem{sandler2018mobilenetv2}
Sandler, M., Howard, A., Zhu, M., Zhmoginov, A., Chen, L.C.: {MobileNetV2}:
  Inverted residuals and linear bottlenecks. In: Proceedings of the IEEE
  Conference on Computer Vision and Pattern Recognition. pp. 4510--4520 (2018)

\bibitem{schroff2015facenet}
Schroff, F., Kalenichenko, D., Philbin, J.: Facenet: A unified embedding for
  face recognition and clustering. In: Proceedings of the IEEE conference on
  computer vision and pattern recognition. pp. 815--823 (2015)

\bibitem{szegedy2015going}
Szegedy, C., Liu, W., Jia, Y., Sermanet, P., Reed, S., Anguelov, D., Erhan, D.,
  Vanhoucke, V., Rabinovich, A.: Going deeper with convolutions. In:
  Proceedings of the IEEE Conference on Computer Vision and Pattern
  Recognition. pp.~1--9 (2015)

\bibitem{whatmough2019fixynn}
Whatmough, P.N., Zhou, C., Hansen, P., Venkataramanaiah, S.K., Seo, J.s.,
  Mattina, M.: {FixyNN}: Efficient hardware for mobile computer vision via
  transfer learning. In: Proceedings of the 2nd SysML Conference (2019)

\bibitem{wu2019trainable}
Wu, C.T., Ain-Kedem, L., Gandra, C.R., Isikdogan, F., Michael, G.: Trainable
  vision scaler (2019), {US Patent App.} 16/232,336

\bibitem{visionisp}
Wu, C.T., Isikdogan, L.F., Rao, S., Nayak, B., Gerasimow, T., Sutic, A., ,
  Ain-kedem, L., Michael, G.: {VisionISP}: Repurposing the image signal
  processor for computer vision applications. In: Proceedings of IEEE
  International Conference on Image Processing (2019)

\bibitem{yosinski2014transferable}
Yosinski, J., Clune, J., Bengio, Y., Lipson, H.: How transferable are features
  in deep neural networks? In: Advances in Neural Information Processing
  Systems. pp. 3320--3328 (2014)

\bibitem{zhang2018shufflenet}
Zhang, X., Zhou, X., Lin, M., Sun, J.: Shufflenet: An extremely efficient
  convolutional neural network for mobile devices. In: Proceedings of the IEEE
  Conference on Computer Vision and Pattern Recognition. pp. 6848--6856 (2018)

\bibitem{zoph2016neural}
Zoph, B., Le, Q.V.: Neural architecture search with reinforcement learning.
  arXiv preprint arXiv:1611.01578  (2016)

\end{thebibliography}
\end{document}